\documentclass[runningheads]{llncs}

\usepackage{todonotes}

\usepackage[T1]{fontenc}
\usepackage{microtype}

\usepackage{amsmath}
\usepackage{amssymb}
\usepackage{stmaryrd}
\usepackage{siunitx}
\DeclareSIUnit\feet{ft}

\usepackage{xcolor}
\usepackage{graphicx}
\usepackage{subcaption}
\usepackage[section]{placeins}

\usepackage{tikz}
\usetikzlibrary{plotmarks}
\usetikzlibrary{arrows.meta}

\usepackage{pgfplots}
\usepgfplotslibrary{groupplots}
\pgfplotsset{compat=1.18}

\usepackage{pgfplotstable}

\usepackage{caption}

\usepackage[frozencache=true, cachedir=minted-cache, newfloat]{minted} 
\setminted{frame=lines}

\usepackage{tabularray}
\UseTblrLibrary{amsmath, booktabs , siunitx}

\usepackage{tblr-extras}
\UseTblrLibrary{caption}

\usepackage{hyperref}
\usepackage[capitalize]{cleveref}

\usepackage{color}

\newcommand{\dl}[2]{\llbracket #1\rrbracket_{\text{#2}}}
\newcommand{\hrect}[1]{\llparenthesis #1\rrparenthesis}

\renewcommand{\implies}{\ \longrightarrow\ }

\DeclareMathOperator*{\argmax}{arg\,max}
\DeclareMathOperator*{\argmin}{arg\,min}

\DeclareMathOperator*{\expected}{\mathbb{E}}

\begin{document}

\title{A General Framework for Property-Driven Machine Learning}

\author{Thomas~Flinkow\inst{1}\orcidID{0000-0002-8075-2194} \and
Marco~Casadio\inst{2}\orcidID{0009-0001-7675-0743} \and
Colin~Kessler\inst{2,3} \and Rosemary~Monahan\inst{1}\orcidID{0000-0003-3886-4675} \and\\ Ekaterina~Komendantskaya\inst{2,4}}

\authorrunning{T. Flinkow et al.}

\institute{Maynooth University, Maynooth, Ireland\\ \email{\{thomas.flinkow,rosemary.monahan\}@mu.ie} \and
Heriot-Watt University, Edinburgh, UK\\ \email{\{mc248,ck2049\}@hw.ac.uk} \and
University of Edinburgh and Edinburgh Centre for Robotics, Edinburgh, UK\\
\and
University of Southampton, Southampton, UK\\ \email{e.komendantskaya@soton.ac.uk}}

\maketitle

\begin{abstract}
Neural networks have been shown to frequently fail to learn critical safety and correctness properties purely from data, highlighting the need for training methods that directly integrate logical specifications.
While adversarial training can be used to improve robustness to small perturbations within $\epsilon$-cubes, domains other than computer vision---such as control systems and natural language processing---may require more flexible input region specifications via generalised hyper-rectangles.
Differentiable logics offer a way to encode arbitrary logical constraints as additional loss terms that guide the learning process towards satisfying these constraints.
In this paper, we investigate how these two complementary approaches can be unified within a single framework for property-driven machine learning, as a step toward effective formal verification of neural networks.
We show that well-known properties from the literature are subcases of this general approach, and we demonstrate its practical effectiveness on a case study involving a neural network controller for a drone system.
Our framework is made publicly available at \url{https://github.com/tflinkow/property-driven-ml}.

\end{abstract}

\section{Introduction}\label{sec:introduction}
Neural networks have been shown to frequently fail to satisfy properties of interest after training~\cite{katzReluplexEfficientSMT2017,casadioNeuralNetworkRobustness2022}, and even the most accurate networks are known to be susceptible to adversarial attacks, i.e. inputs that resemble the training data but fall outside of the learnt distribution~\cite{szegedyIntriguingPropertiesNeural2014,goodfellowExplainingHarnessingAdversarial2015}.
This puts restrictions on their use in safety-critical domains.
To address this gap, numerous formal verification tools\footnote{For an overview of the state-of-the art in formal verification of neural networks, we refer the interested reader to~\cite{huangSurveySafetyTrustworthiness2020,urbanReviewFormalMethods2021,liuAlgorithmsVerifyingDeep2021,albarghouthiIntroductionNeuralNetwork2021}.} have been proposed in the past few years, including Marabou~\cite{katzReluplexEfficientSMT2017,katzMarabouFrameworkVerification2019,wuMarabouVersatileFormal2024}, Branch-and-Bound~\cite{bunelBranchBoundPiecewise2020}, NNV~\cite{tranNNVNeuralNetwork2020,lopezNNV20Neural2023}, and $\alpha,\beta$-CROWN~\cite{zhangEfficientNeuralNetwork2018a,xuAutomaticPerturbationAnalysis2020a,xuFastCompleteEnabling2021,wangBetaCROWNEfficientBound2021,zhangGeneralCuttingPlanes2022a,shiNeuralNetworkVerification2024} (winner of the recent Neural Network Verification Competitions (VNN-COMP)~\cite{bakSecondInternationalVerification2021,mullerThirdInternationalVerification2022,brixFourthInternationalVerification2023,brixFirstThreeYears2023b}).

Clearly, formal verification is only effective if the trained networks satisfy the desired properties---which standard data-driven training often fails to ensure.
This motivates the need to integrate data- and property-driven training.

\paragraph{Standard optimisation methods that improve robustness.}
In the simplest case, data augmentation and adversarial training~\cite{goodfellowExplainingHarnessingAdversarial2015,madryDeepLearningModels2018} can be used to improve a network's robustness.
Data augmentation artificially enlarges the data set (e.g. via noise, rotation, etc.).
Adversarial training aims to find the worst-case perturbation within $\epsilon$-distance around an input and integrates the loss computed for the perturbation into the training process.
As an example, consider a property frequently used in image classification: \emph{local robustness} of a neural network $f$, expressed as $\forall\vec{x}'\ldotp\lVert\vec{x}-\vec{x}'\rVert\le\epsilon\implies
\lVert f(\vec{x})-f(\vec{x}')\rVert\le\delta$.
Given an image $\vec{x}$ in the data set, the network's prediction may deviate by at most $\delta$ when perturbing an input image $\vec{x}$ by at most $\epsilon$.
Adversarial training (e.g. via Fast Gradient Sign Method (FGSM~\cite{goodfellowExplainingHarnessingAdversarial2015}) or Projected Gradient Descent (PGD)~\cite{madryDeepLearningModels2018}) finds the perturbation in the $\epsilon$-cube around $\vec{x}$ for which the property $\lVert f(\vec{x})-f(\vec{x}')\rVert \le \delta$ fails and optimises the network towards correctly classifying that sample.

Although these standard optimisation methods generally work well for computer vision models, they fail to generalise beyond this domain.
In many scenarios we encounter properties of the form $\forall\vec{x}\ldotp\mathcal{P}(\vec{x})\implies 
\mathcal{Q}(f(\vec{x}))$, where preconditions $\mathcal{P}$ and postconditions $\mathcal{Q}$ differ substantially from the concrete instances used in the algorithms that optimise networks for local robustness.
First, properties $\mathcal{P}$ and $\mathcal{Q}$ do not have to relate to specific data points.
Global properties of the input space can constrain input vectors in arbitrary ways. 
Secondly, $\mathcal{P}$ and $\mathcal{Q}$ do not have to be atomic, and can be given by arbitrary logical formulas.

Recently, several methods have been proposed in order to generalise adversarial training to optimisation tasks involving global and non-atomic properties.

\paragraph{Global properties and hyper-rectangles.}
Generally, for a neural network $f\colon\mathbb{R}^m\to\mathbb{R}^n$, a global property $\mathcal{P}$ on the input space $\mathbb{R}^m$ is given by a number of constraints of the form $l_i\le x_i\le u_i$, for constants $l_i$, $u_i$, and $\vec{x}=[x_1,\ldots,x_m]\in\mathbb{R}^m$ being an element of the input space.
Such constraints generally give rise to the notion of a \emph{hyper-rectangle} (described in more detail in~\cref{subsec:input_region_spacification}), which generalises the notion of $\epsilon$-cubes frequently deployed in adversarial training~\cite{casadioANTONIOSystematicMethodb,casadioNLPVerificationGeneral2025a,floodFormallyVerifyingRobustness2025}.

Although initially investigated in the context of classifiers for large language models~\cite{casadioANTONIOSystematicMethodb,casadioNLPVerificationGeneral2025a} and for security applications~\cite{floodFormallyVerifyingRobustness2025}, there is an even more common application domain for such properties: neural networks deployed in cyber-physical systems as \emph{neural controllers}.
The seminal neural network verification benchmark \emph{ACAS Xu}\footnote{The experimental neural network compression~\cite{julianPolicyCompressionAircraft2016} of the unmanned variant \emph{ACAS Xu} of the \emph{Airbourne Collision Avoidance System X (ACAS X)}~\cite{kochenderferRobustAirborneCollision2011,kochenderferOptimizedAirborneCollision2015}.}, for example, is defined via ten global properties that constrain sensor readings based on their intended semantics such as velocity, distance and angle to the intruder.
Many neural controllers yield similar verification conditions~\cite{manzanaslopezARCHCOMP24CategoryReport2024}.

\paragraph{Non-atomic postconditions and differentiable logics.}
Important examples where the postcondition $\mathcal{Q}$ is non-atomic have also been flagged in the literature~\cite{seshiaFormalSpecificationDeep2018a,farrellExploringRequirementsSoftware2023}.
Often, constraints on the output should not just encourage the network to make the correct class prediction, but rather encode complex, expressive constraints on the outputs: for example, in DL2~\cite{fischerDL2TrainingQuerying2019}, a constraint for CIFAR-100~\cite{krizhevskyLearningMultipleLayers2009} that refers to the probability of a group of people (consisting of the individual probabilities of the classes for baby, boy, girl, man, and woman) can be used to make misclassifications less severe; the network can be taught to make it more preferable to misclassify a man as a boy rather than as a bicycle.

It was suggested that \emph{differentiable logics} (described in more detail in~\cref{subsec:property_driven_training}) can be used to translate arbitrary logical constraints into additional loss terms for optimisation.
These loss terms generalise the loss $\lVert f(\vec{x})-f(\vec{x}')\rVert \le \delta$ used in the PGD or FGSM optimisation algorithms to a function that encodes an arbitrary property $\mathcal{Q}(f(\vec{x}))$.
This line of research was explored in~\cite{fischerDL2TrainingQuerying2019,vankriekenAnalyzingDifferentiableFuzzy2022,flinkowComparingDifferentiableLogics2025}.

\paragraph{Contributions.}
This paper presents a \emph{generalised methodology} for property-driven machine learning, combining the two existing lines of research into training with hyper-rectangles and differentiable logics.
The result is a method that optimises arbitrary neural networks to arbitrary properties $\forall\vec{x}\ldotp\mathcal{P}(\vec{x})\implies 
\mathcal{Q}(f(\vec{x}))$ written in a subset of first-order logic.
In particular, $\mathcal{P}(\vec{x})$ and $\mathcal{Q}(f(\vec{x}))$ can be expressed by arbitrary quantifier-free first-order formulae, such that occurrences of the neural network $f$ is permitted in $\mathcal{Q}$ but not in $\mathcal{P}$, and  $\mathcal{P}$ is limited to constraints that have constant upper and lower bounds for each dimension.

We argue that this class of generalised properties is characteristic of properties deployed in verification of neural network controllers used in cyber-physical systems. This happens for two reasons:
\begin{itemize}
    \item neural controller modelling often comes with well-defined global constraints concerning input regions.
    These are usually given as assumptions on safe and unsafe sensor readings that, unlike pixels in images, have well-defined mathematical or physical meaning.
    This can give rise to a non-trivial property $\mathcal{P}$ expressed as a complex Boolean formula;
    \item a degree of freedom in the behaviour of a neural controller is expected: i.e. we do not want to firmly prescribe the controller in every scenario, but want to only constrain unsafe actions.
    For example, for ACAS Xu, the property $\mathcal{P}$ \emph{``the distance to intruder is dangerously small''} does not mandate \emph{``advise `strong left'{}''}, but is rather expressed as \emph{``do not advise `clear of conflict'{}''}, which boils down to a disjunctive prescription for $\mathcal{Q}$: \emph{``advise `weak left', `strong left', `weak right', or `strong right'{}''}.
    Generally, such verification scenarios result in non-trivial $\mathcal{Q}$. 
\end{itemize}
We deploy a new case-study of a neural network controller (see~\cite{kesslerNeuralNetworkVerification2025} for more details) and use it to test our method.
This use case not only gives us more flexibility in both property specification and training (as the ACAS Xu data set is not publicly released), but also presents a generalisation of the ACAS Xu case to regression models. 

\paragraph{Paper outline.}
In the following, we will first set the context of this work, explaining the theory and gaps of input region specification in~\cref{subsec:input_region_spacification}, and of differentiable logics in~\cref{subsec:property_driven_training}, respectively.
We describe our unifying framework combining (1) flexible input region specification via hyper-rectangles and (2) differentiable logics for translating arbitrary logical constraints into additional loss terms in~\cref{sec:combined_framework} and show how well-known properties from the literature emerge as special cases of this framework.
We briefly describe our Python implementation, publicly available at \url{https://github.com/tflinkow/property-driven-ml}.
We show the effectiveness of this approach on MNIST and a case study involving a neural network controller for a glider drone in~\cref{sec:experimental_setup}.
Lastly, we conclude with a discussion, including limitations and possible areas of future work, in~\cref{sec:discussion}.

\section{Context}

\subsection{Input Region Specification}\label{subsec:input_region_spacification}
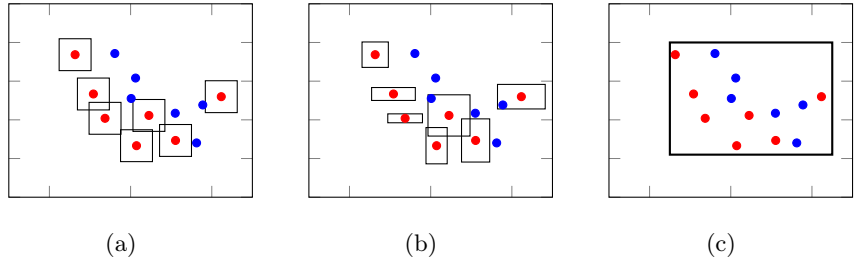
\begin{figure}
    \centering
    \begin{subfigure}[t]{0.3\textwidth}
    \centering
    \begin{tikzpicture}
        \begin{axis}[
            width=4.8cm,
            xticklabels=\empty, yticklabels=\empty,
            xmin=-1, xmax=5, ymin=-2, ymax=8,
        ]
        
        \pgfmathsetseed{45}
        
        \foreach \i in {1,...,9} {
            \pgfmathsetmacro\x{rnd*5 + 0.5}
            \pgfmathsetmacro\y{rnd*5 + 0.5}
            
            \addplot[only marks, mark=square, draw=black, fill=none, mark size=6pt] coordinates {(\x,\y)};
            
            \addplot[only marks, mark=*, mark options={fill=red, draw=red}, mark size=1.5pt] coordinates {(\x,\y)};
        }
        
        \foreach \i in {1,...,6} {
            \pgfmathsetmacro\x{rnd*5 + 0.5}
            \pgfmathsetmacro\y{rnd*5 + 0.5}
        
            \addplot[only marks, mark=*, mark options={fill=blue, draw=blue}, mark size=1.5pt] coordinates {(\x,\y)};
        }
        
        \end{axis}
    \end{tikzpicture}
    \caption{}
    \label{fig:epsilon_balls}
\end{subfigure}%
\hfil
\begin{subfigure}[t]{0.3\textwidth}
    \centering
    \begin{tikzpicture}
        \begin{axis}[
            width=4.8cm,
            xticklabels=\empty, yticklabels=\empty,
            xmin=-1, xmax=5, ymin=-2, ymax=8,
        ]
        
        \pgfmathsetseed{45}


        \foreach \i in {1,...,9} {
            \pgfmathsetmacro{\xval}{rnd*5 + 0.5}
            \pgfmathsetmacro{\yval}{rnd*5 + 0.5}
            \expandafter\xdef\csname redx\i\endcsname{\xval}
            \expandafter\xdef\csname redy\i\endcsname{\yval}
        }

        \pgfmathsetseed{15}
        
        \foreach \i in {1,...,9} {
            \pgfmathsetmacro{\x}{\csname redx\i\endcsname}
            \pgfmathsetmacro{\y}{\csname redy\i\endcsname}
            
            \pgfmathsetmacro{\sx}{rnd*1.5}
            \pgfmathsetmacro{\sy}{rnd*1.5}

            \edef\temp{
                \noexpand\addplot[
                    only marks,
                    mark=square,
                    draw=black,
                    fill=none,
                    mark size=6pt,
                    mark options={xscale=\sx, yscale=\sy}
                ] coordinates {(\x,\y)};
            }
            \temp
        
            \addplot[
                only marks,
                mark=*,
                mark options={fill=red, draw=red},
                mark size=1.5pt
            ] coordinates {(\x,\y)};
        }
        
        \pgfmathsetseed{45}
        \foreach \i in {1,...,9} {
            \pgfmathparse{rnd} 
            \pgfmathparse{rnd} 
        }
        
        \foreach \i in {1,...,6} {
            \pgfmathsetmacro\x{rnd*5 + 0.5}
            \pgfmathsetmacro\y{rnd*5 + 0.5}
        
            \addplot[only marks, mark=*, mark options={fill=blue, draw=blue}, mark size=1.5pt]
                coordinates {(\x,\y)};
        }            
        \end{axis}
    \end{tikzpicture}
    \caption{}
    \label{fig:local_hyper-rectangles}
\end{subfigure}%
\hfil
\begin{subfigure}[t]{0.3\textwidth}
    \centering
    \begin{tikzpicture}
        \begin{axis}[
            width=4.8cm,
            xticklabels=\empty, yticklabels=\empty,
            xmin=-1, xmax=5, ymin=-2, ymax=8,
        ]
        
        \pgfmathsetseed{45}
        
        \foreach \i in {1,...,9} {
            \pgfmathsetmacro\x{rnd*5 + 0.5}
            \pgfmathsetmacro\y{rnd*5 + 0.5}
            
            \addplot[only marks, mark=*, mark options={fill=red, draw=red}, mark size=1.5pt] coordinates {(\x,\y)};
        }
        
        \foreach \i in {1,...,6} {
            \pgfmathsetmacro\x{rnd*5 + 0.5}
            \pgfmathsetmacro\y{rnd*5 + 0.5}
        
            \addplot[only marks, mark=*, mark options={fill=blue, draw=blue}, mark size=1.5pt] coordinates {(\x,\y)};
        }
        
        \draw[thick] (axis cs:0.5,0.2) rectangle (axis cs:4.5,6);
        
        \end{axis}
    \end{tikzpicture}
    \caption{}
    \label{fig:global_hyper-rectangles}
\end{subfigure}%
%
    \caption{A visualisation of $\epsilon$-cubes (\cref{fig:epsilon_balls}), hyper-rectangles relative to each data point (\cref{fig:local_hyper-rectangles}), and a global hyper-rectangle (\cref{fig:global_hyper-rectangles}). Red dots represent training data, and blue dots represent test data.}
    \label{fig:hyper-rectangles}
\end{figure}
\noindent As noted in~\cite{casadioNLPVerificationGeneral2025a}, verification of neural networks has focused primarily on \emph{robustness verification}, which requires a classification network $f\colon\mathbb{R}^m\to\mathbb{R}^n$ to assign the same class label for each element in each subspace $\mathcal{S}_i\subseteq\mathbb{R}^m$.
For computer vision tasks, these subspaces often take the shape of $\epsilon$-balls around all input vectors in the dataset.
Formally, given an input $\vec{x}$ and distance function $\lVert\cdot\rVert$, the \emph{$\epsilon$-ball} around $\vec{x}$ of radius $\epsilon$ is defined as:
\begin{equation}\label{eq:eps_cube}
    \mathbb{B}(\vec{x};\epsilon):=\{\vec{x}'\in\mathbb{R}^m: \lVert\vec{x}-\vec{x}'\rVert\le\epsilon\}
\end{equation}
In practice, it is common to use the $\ell_{\infty}$ norm~\cite{goodfellowExplainingHarnessingAdversarial2015}, in which case these $\epsilon$-balls are also called $\epsilon$-cubes.

While $\epsilon$-bounded perturbations work reasonably well for computer vision tasks~\cite{wongProvableDefensesAdversarial2018,gowalScalableVerifiedTraining2019}, they are not necessarily as well suited for other tasks, such as natural language processing or low-dimensional problems (e.g. neural network controllers).
In computer vision, images are seen as vectors in a continuous space and thus every point in that space corresponds to a valid image. In contrast, natural language processing input spaces are usually discrete, and $\epsilon$-balls around one sentence embedding may not in fact contain any other valid sentences~\cite{casadioNLPVerificationGeneral2025a}.

Although the idea of generalising $\epsilon$-cubes to arbitrary hyper-rectangles has existed for a while, Casadio et al.~\cite{casadioNLPVerificationGeneral2025a} were the first to argue that hyper-rectangles can systematically be used in verification and property-driven training of neural networks.
A multi-dimensional rectangle (or \emph{hyper-rectangle}), is a shape defined by lower and upper bounds for each dimension, i.e. $\vec{l},\vec{u}\in\mathbb{R}^m$ such that $\vec{l}\le\vec{u}$ (component-wise).
The hyper-rectangle $H(\vec{l},\vec{u})$ is then defined as
\begin{equation}\label{eq:hyper-rectangle}
    H(\vec{l},\vec{u}):=\{\vec{x}\in\mathbb{R}^m \mid l_i\le x_i\le u_i,\ \text{for all }1\le i\le m\}.
\end{equation}
As visualised in~\cref{fig:hyper-rectangles}, hyper-rectangles allow the specification of $\epsilon$-cubes, as well as more flexible local regions, and they can also be used to specify global input regions (not relative to specific inputs)\footnote{For examples of such properties see~\cite{floodFormallyVerifyingRobustness2025,casadioANTONIOSystematicMethodb}.}.

\subsection{Property-driven Training}\label{subsec:property_driven_training}
\paragraph{Training for robustness.}
In standard machine learning, for a model $f_{\theta}$ parametrised by $\theta$, optimal parameters $\theta^*$ are obtained by using gradient descent to minimise a \emph{loss} $\mathcal{L}$ that quantifies the difference between the network's actual output $f_{\theta}(\vec{x})$ and the desired output $\vec{y}$, shown in~\cref{eq:normal_ml}:
\begin{equation}\label{eq:normal_ml}
    \theta^*=\argmin\limits_{\theta}\expected\limits_{(\vec{x},\vec{y})\sim\mathcal{D}}\ \bigl[\mathcal{L}(\vec{x},\vec{y};f_{\theta})\bigr]
\end{equation}
Adversarial training via FGSM and PGD algorithms~\cite{goodfellowExplainingHarnessingAdversarial2015,madryDeepLearningModels2018} first finds adversarial examples and integrates them into the training process.
Formally, the minimisation objective changes from the one shown in~\cref{eq:normal_ml} to the one shown in~\cref{eq:min_max_ml}, where $\vec{x}'$ is an adversarial example within the $\epsilon$-cube around a given input $\vec{x}$.
\begin{equation}\label{eq:min_max_ml}
    \theta^*=\argmin\limits_{\theta}\expected\limits_{(\vec{x},\vec{y})\sim\mathcal{D}}\ \biggl[\max\limits_{\vec{x}'\in\mathbb{B}(\vec{x};\epsilon)}\mathcal{L}(\vec{x}',\vec{y};f_{\theta})\biggr]
\end{equation}
Therefore, adversarial training consists of a two-step process: an inner maximisation problem to select the worst-case adversarial perturbation within the $\epsilon$-cube around a clean data point $\vec{x}$, followed by adapting the network weights to reduce the loss caused by this adversarial perturbation.
Madry et al.~\cite{madryDeepLearningModels2018} proposed to use PGD to approximately solve the inner optimisation problem.

\paragraph{Training to satisfy arbitrary logical constraints.}
Adversarial training aims to improve prediction accuracy on perturbed inputs, i.e. similar inputs should be assigned the same class label.
It has been shown in~\cite{casadioNeuralNetworkRobustness2022} that training for one kind of robustness does not necessarily imply another; for example, while adversarial training improves local robustness (i.e. given an input $\vec{x}$, standard robustness requires $\forall\vec{x}'\in\mathbb{B}(\vec{x};\epsilon)\ldotp\lVert f(\vec{x})-f(\vec{x}')\rVert\le\delta$), most works on verification of neural networks focus on classification robustness (i.e., given an input $\vec{x}$ and true label $y$, classification robustness requires $\forall\vec{x}'\in\mathbb{B}(\vec{x};\epsilon)\ldotp\argmax_i f(\vec{x}')=y$).

So-called \emph{differentiable logics} have been proposed to generate additional loss functions from arbitrary logical constraints, not limited to improving prediction accuracy.
This additional loss term calculates a penalty depending on how much the network deviates from satisfying the constraint.
Popular choices for differentiable logics include both systems designed specifically for use in neural networks, such as \emph{Deep Learning with Differentiable Logics (DL2)}~\cite{fischerDL2TrainingQuerying2019}, as well as already existing and well-studied logical systems such as \emph{fuzzy logics}.
\begin{table}[tb]
    \centerline{%
        \begin{talltblr}
        [
            caption={An overview of some popular differentiable logics implemented in our framework. $\dl{\cdot}{}$ denotes the mapping of a logical formula into loss.},
            label={tab:differentiable_logics},
            note{a}={Atoms in DL2 are $\dl{x\le y}{DL2}$ and $\dl{x\neq y}{DL2}$. Instead of providing a dedicated negation operator, negation is pushed inwards to the level of comparison.},
            note{b}={In~\cite{varnaiRobustnessMetricsLearning2020}, an $n$-ary conjunction operator is proposed that satisfies desirable properties (such as commutativity and shadow-lifting). We omit its definition here for brevity.}
        ]
        {
            colspec={Q[l, m, mode=text]Q[c, m, mode=math]Q[c, m, mode=math]Q[c, m, mode=math]Q[c, m, mode=math]Q[c, m, mode=math]Q[c, m, mode=math]},
            row{1}={mode=text, font=\bfseries},
        }
            \toprule
                Logic & Domain & $\dl{\top}{}$ & $\dl{\bot}{}$ & $\dl{\lnot x}{}$ & $\dl{x\wedge y}{}$ & $\dl{x\vee y}{}$ \\
            \midrule
                DL2~\cite{fischerDL2TrainingQuerying2019} & [0, \infty) & 0 & \infty & \text{undefined}\TblrNote{a} & xy & x+y \\
                Gödel logic~\cite{vankriekenAnalyzingDifferentiableFuzzy2022} & [0, 1] & 1 & 0 & 1-x & \min\{x,y\} & \max\{x,y\} \\
                Product logic~\cite{vankriekenAnalyzingDifferentiableFuzzy2022} & [0, 1] & 1 & 0 & 1-x & xy & x+y-xy \\
                STL~\cite{varnaiRobustnessMetricsLearning2020} & (-\infty,\infty) & \infty & -\infty & -x & \text{omitted}\TblrNote{b} & \text{omitted}\TblrNote{b} \\ 
            \bottomrule
        \end{talltblr}
    }
\end{table}
The first in-depth study investigating the use of fuzzy logics as loss functions (called \emph{Differentiable Fuzzy Logics}) was provided by \cite{vankriekenAnalyzingDifferentiableFuzzy2022} and presented insights into the learning characteristics of differentiable logic operators.
A unifying framework of differentiable logics including DL2, fuzzy logics, and Signal Temporal Logic (STL)~\cite{varnaiRobustnessMetricsLearning2020}, called \emph{Logic of Differentiable Logics (LDL)}, is provided in~\cite{slusarzLogicDifferentiableLogics2023a} and allows for mechanised proofs of theoretical properties of differentiable logics~\cite{affeldtTamingDifferentiableLogics2024}.

There are many differentiable logics to choose from, and these differ not only in terms of their domains and operators (compare~\cref{tab:differentiable_logics}), but also when it comes to satisfying properties such as soundness, compositionality, smoothness, and shadow-lifting (which requires the truth value of a conjunction to increase whenever the truth value of one of its conjuncts increases).
It has been shown~\cite{affeldtTamingDifferentiableLogics2024,varnaiRobustnessMetricsLearning2020} that no differentiable logic can satisfy all desirable properties.

Recently, a general differentiable logic library was introduced in~\cite{flinkowComparingDifferentiableLogics2025}, and implements major known differentiable logics under the same umbrella, allowing for their empirical comparative evaluation for the first time.
Therein, it is assumed that all constraints are of the form $\forall\vec{x}'\in\mathbb{B}(\vec{x};\epsilon)\ldotp\phi$, i.e. a logical constraint should not only hold for a specific input $\vec{x}$, but also in its close vicinity.
Therefore, the optimisation objective for the constraint $\phi$ is shown in~\cref{eq:old_optimisation}:
\begin{equation}\label{eq:old_optimisation}
    \theta^*=\argmin\limits_{\theta}\expected\limits_{(\vec{x},\vec{y})\sim\mathcal{D}}\ \biggl[\lambda\mathcal{L}(\vec{x},\vec{y};f_{\theta})+(1-\lambda)\max\limits_{\vec{x}'\in\mathbb{B}(\vec{x};\epsilon)} \dl{\phi}{}(\vec{x},\vec{x}',\vec{y};f_{\theta})\biggr].
\end{equation}
Here, $\mathcal{L}$ denotes the prediction loss\footnote{Such as cross-entropy loss $\mathcal{L}_{\text{CE}}$ for classification, mean-squared-error loss $\mathcal{L}_{\text{MSE}}$ for regression, etc.}, 
$(\vec{x},\vec{y})\sim\mathcal{D}$ is a pair of training data $\vec{x}$ and target $\vec{y}$ (i.e. the true label for classification, or the target value for regression), and $\lambda$ is a hyperparameter required to balance the different loss terms.

In this paper, we will extend that library to cover property-driven training with general properties of the form $\forall\vec{x}\ldotp\mathcal{P}(\vec{x})\implies 
\mathcal{Q}(f(\vec{x}))$ (subject to the limitations described in~\cref{sec:introduction}).
At the same time, we inherit the functionality of compiling $\mathcal{Q}(f(\vec{x}))$ to various differentiable logics, including DL2, fuzzy-logic based ones, and STL.

\section{Combining Input Region Specification and Logical Constraint-based Loss}\label{sec:combined_framework}

\subsection{Definition of the Novel Optimisation Objective}
We now introduce our generalised method formally. Given: 
\begin{itemize}
    \item a neural network $f_{\theta}\colon\mathbb{R}^m\to\mathbb{R}^n$ parametrised by training parameters $\theta$ (such as neural network weights), and 
    \item a constraint $\phi\colon\forall\vec{x}\in\mathbb{R}^m\ldotp \mathcal{P}(\vec{x})\implies \mathcal{Q}(f(\vec{x}))$ which we want to incorporate into optimisation for $\theta$,
\end{itemize}
we define a translation function $\hrect{\cdot}$ from a precondition $\mathcal{P}(\vec{x})$ into a hyper-rectangle as follows.
Assume $\mathcal{P}(\vec{x})$ is given by an arbitrary Boolean formula containing atomic propositions  only of the form $l_i\le x_i\le u_i$, for constants $l_i$, $u_i$, and $\vec{x}=[x_1,\ldots,x_m]\in\mathbb{R}^m$ is a list of variables that denotes an input vector.
Let $\vec{l}^{\text{min}}$ and $\vec{u}^{\text{max}}$ be constant vectors that define minimum lower and maximum upper bounds of $\vec{x}$\footnote{In neural network verification, we always assume default lower and upper bounds are inherently present (inferred from data) when some element vectors are not explicitly constrained. Examples in the rest of the paper illustrate that point.}. 

If $\mathcal{P}(\vec{x})$ is not satisfiable, we define $\hrect{\mathcal{P}(\vec{x})}=\emptyset$.
Otherwise, we define $\hrect{\mathcal{P}(\vec{x})}$ by induction on the structure of $\mathcal{P}(\vec{x})$:
\begin{itemize}
    \item If $\mathcal{P}(\vec{x})$ is of the form $l_i\le x_i\le u_i$, we form a hyper-rectangle $H(\vec{l}^*,\vec{u}^*)$, where $\vec{l}^* = \vec{l}^{\text{min}}$ except for its $i$th component being $l_i$ (and similarly obtain $\vec{u}^*$), and define $\hrect{\mathcal{P}(\vec{x})}=H(\vec{l}^*,\vec{u}^*)$;
    \item if $\mathcal{P}(\vec{x})$ is of the form $\lnot F$, where $F$ is an arbitrary formula, then  $\hrect{F}=\hrect{F}^{\sim}$;
    \item if $\mathcal{P}(\vec{x})$ is of the form $F_1\vee F_2$, where $F_1, F_2$ are arbitrary formulas, then $\hrect{F_1\vee F_2}=\hrect{F_1}\cup\hrect{F_2}$;
    \item and lastly, if $\mathcal{P}(\vec{x})$ is of the form $F_1\wedge F_2$, then $\hrect{F_1\wedge F_2}=\hrect{F_1}\cap\hrect{F_2}$;
\end{itemize}
where $^{\sim}, \cup, \cap$ stand for set-theoretic complement, union, and intersection. 

Recall that, by the construction of the previous section, we next define a translation function $\dl{\cdot}{}$ from the postcondition $\mathcal{Q}(f(\vec{x}))$ into real-valued loss, thus $\dl{\mathcal{Q}(f(\vec{x}))}{}$ is defined.

\noindent Then, the optimisation objective for the constraint $\phi$ is given as follows:
\begin{equation}\label{eq:combined}
    \theta^*=\argmin\limits_{\theta}\expected\limits_{(\vec{x},\vec{y})\sim\mathcal{D}}\ \biggl[\lambda\mathcal{L}(\vec{x},\vec{y};f_{\theta})+(1-\lambda)\max\limits_{\vec{x}'\in\hrect{\mathcal{P}(\vec{x})}} \dl{\mathcal{Q}(f(\vec{x}))}{}(\vec{x},\vec{x}',\vec{y};f_{\theta})\biggr].
\end{equation}
Note the similarity to the optimisation objective previously shown in~\cref{eq:old_optimisation}. However, to find adversarial examples for the constraint loss, the inner maximisation problem $\max_{\vec{x}'\in\hrect{\mathcal{P}(\vec{x})}} \dl{\mathcal{Q}(f(\vec{x}))}{}(\vec{x},\vec{x}',\vec{y};f_{\theta})$ is approximated using a modified PGD attack that works with arbitrary hyper-rectangles $\hrect{\mathcal{P}(\vec{x})}$.

\subsection{Examples from the Literature in Our Terms}
We briefly show how well-studied properties from the literature can be seen as sub-cases of the unified approach we proposed in this section.

\paragraph{ACAS Xu.}
The ACAS Xu neural networks take six sensor measurements as inputs and produce scores for five possible advisories, ``clear-of-conflict'' (COC), ``weak left'' (WL), ``weak right'' (WR), ``strong left'' (SL), or ``strong right'' (SR).

As a representative example, we take property $\phi_2$ from~\cite{katzReluplexEfficientSMT2017arxiv}, which requires that the score for a ``clear of conflict'' advisory should never be maximal for a distant, significantly slower intruder, meaning that it can never be the worst action to take.
\begin{itemize}
    \item\textbf{Obtaining $\hrect{\mathcal{P}}$:} The \emph{intruder being distant and significantly slower} translates to the following lower bounds on network inputs:
    \begin{equation}\label{eq:acasxu_input_bounds}
        \qty{55947.691}{\feet}\le\rho, \quad \qty{1145}{\feet\per\second}\le v_{\text{own}}, \quad v_{\text{intruder}}\le\qty{60}{\feet\per\second},
    \end{equation}
    which directly give rise to a hyper-rectangle:
    assuming the inputs to the network to be a vector $\vec{x}=[\rho,\theta,\psi,v_{\text{own}},v_{\text{intruder}}]$, the hyper-rectangle induced by~\cref{eq:acasxu_input_bounds} translates to lower bound $\vec{l}$ and upper bound $\vec{u}$ as
    \begin{equation}
        \vec{l}=\max\{\vec{l}^{\text{min}},[\qty{55947.691}{\feet},-\infty,-\infty,\qty{1145}{\feet\per\second},\qty{60}{\feet\per\second}]\},\ \text{and}\ \vec{u}=\vec{u}^{\text{max}},
    \end{equation}
    where $\vec{l}^{\text{min}}$ and $\vec{u}^{\text{max}}$ denote the minimum and maximum permitted values for each of the six parameters, and $\max\{\cdot,\cdot\}$ is applied component-wise.
    \item\textbf{Obtaining $\dl{\mathcal{Q}(f(\vec{x}))}{}$:} The constraint on the output, i.e. \emph{the score for a ``clear of conflict'' advisory should never be maximal}, can be expressed as the disjunction
    \begin{equation}
        (y_{\text{COC}}<y_{\text{WL}})\vee(y_{\text{COC}}<y_{\text{WR}})\vee(y_{\text{COC}}<y_{\text{SL}})\vee(y_{\text{COC}}<y_{\text{SR}}),
    \end{equation}
    and translated into loss using the mappings $\dl{t<t'}{}$ and $\dl{t\vee t'}{}$ of a chosen differentiable logic.
    The actual translation is omitted here for the sake of brevity,
    but can be obtained by repeated application of $\dl{t\vee t'}{}$ and the identity $\dl{t<t'}{}=\dl{t\le t'\wedge t\neq t'}{}$.
\end{itemize}

\paragraph{Standard robustness.}
A property that is often of interest for image classification problems is \emph{standard robustness}~\cite{casadioNeuralNetworkRobustness2022}, expressed as
\begin{equation}\label{eq:standard_robustness}
    \forall\vec{x}'\in\mathbb{B}(\vec{x};\epsilon)\ldotp\lVert f(\vec{x})-f(\vec{x}')\rVert\le\delta.
\end{equation}
This property states that there is a maximum allowed distance the output may be perturbed for a slightly perturbed input.

\begin{itemize}
    \item\textbf{Obtaining $\hrect{\mathcal{P}(\vec{x})}$:} In this case, the input region specification is obtained by combining the $\epsilon$-cubes around each image $\vec{x}_i$ in the training set $\{(\vec{x}_i,\vec{y}_i)\}_{i=1}^n$ and filtering on only valid images\footnote{Without loss of generality, assume that pixels can take a value between $0$ and $1$, and thus valid images made up of $m$ pixels are in the set $[0,1]^m$.}, thus we obtain
    \begin{equation}
        \hrect{\mathcal{P}(\vec{x})}=[0,1]^m\cap\mathbb{B}(\vec{x};\epsilon).
    \end{equation}
    \item\textbf{Obtaining $\dl{\mathcal{Q}(f(\vec{x}))}{}$:} The output constraint is obtained using the mapping $\dl{t\le t'}{}$ of a chosen differentiable logic.
    For example, using DL2's mapping $\dl{t\le t'}{DL2}=\max\{0, t-t'\}$, the resulting loss function would be
    \begin{equation}
        \dl{\mathcal{Q}(f(\vec{x}))}{DL2}=\max\{0, \lVert f(\vec{x})-f(\vec{x}')\rVert - \delta\}.
    \end{equation}
\end{itemize}

\subsection{Implementation}
We briefly describe our PyTorch~\cite{paszkePyTorchImperativeStyle2019} implementation of this training framework, publicly available at \url{https://github.com/tflinkow/property-driven-ml}.
Our main aim is to provide a generic, reusable framework that is easy to use and easy to adapt to new scenarios, while at the same time is 
computationally efficient.

Hyper-rectangles are internally stored as lower and upper bounds of the same shape as the data, allowing for efficient operations (such as sampling and projection) for PGD.
For convenience, shortcuts such as \mintinline{python}{EpsilonBall(eps: float)} are implemented, which automatically create upper and lower bounds for each data point in the dataset.
Our implementation makes these hyper-rectangles part of the dataset; by extending \mintinline{python}{torch.utils.data.Dataset}, the corresponding hyper-rectangles can easily be obtained with the training data and labels.
\Cref{lst:dataloader} shows how to create and use $\epsilon$-balls for MNIST during training.

We take on a parametric approach for expressing logical constraints: an abstract \mintinline{python}{Logic} object provides common operators such as \mintinline{python}{LEQ(x, y)}, \mintinline{python}{NOT(x)}, $n$-ary \mintinline{python}{AND(*xs)} and \mintinline{python}{OR(*xs)}, \mintinline{python}{IMPLIES(x, y)}, and \mintinline{python}{EQUIV(x, y)}, which allows the user to specify constraints (see~\cref{lst:loss_implementation}) without having to choose a specific differentiable logic or to even be aware of differences in their operators or domains.

\begin{listing}
    \caption{Constructing and using $\epsilon$-balls during training. A standard \mintinline{python}{DataLoader} automatically obtains the corresponding hyper-rectangles in the form of lower and upper bounds (\mintinline{python}{lo} and \mintinline{python}{hi}) for each input \mintinline{python}{image} in each batch.}
    \label{lst:dataloader}\vspace{-0.6cm}
\begin{minted}{python}
dataset = torchvision.datasets.MNIST(train=True)
dataset = EpsilonBall(dataset, eps=0.3)

loader = torch.utils.data.DataLoader(dataset)

# obtain image, label, and corresponding lower and upper bounds
for _, (image, label, lo, hi) in enumerate(loader):
    ...
\end{minted}
\end{listing}
\begin{listing}
    \caption{An implementation of a Lipschitz robustness constraint, specifying $\lVert f(\vec{x})-f(\vec{x}_{\text{adv}})\rVert_2\le L\lVert\vec{x}-\vec{x}_{\text{adv}}\rVert_2$. Logical operations are specified using an abstract logic \mintinline{python}{l} which provides operations such as \mintinline{python}{LEQ(x, y)} and can later be instantiated with various differentiable logics, allowing for separation of property specification and translation into real-valued operations.}
    \label{lst:loss_implementation}\vspace{-0.6cm}
\begin{minted}{python}
class LipschitzRobustnessConstraint(Constraint):
    def get_constraint(self, N, x, x_adv, y_target):
        y, y_adv = N(x), N(x_adv)

        diff_x = torch.linalg.vector_norm(x_adv - x, ord=2, dim=1)
        diff_y = torch.linalg.vector_norm(y_adv - y, ord=2, dim=1)

        return lambda l: l.LEQ(diff_y, L * diff_x)
\end{minted}
\end{listing}

Lastly, we provide two important enhancements that reduce the number of hyperparameters the user has to tune, while at the same time achieving higher effectiveness of the training process: (1) balancing the two loss terms is handled by the adaptive loss-balancing algorithm GradNorm~\cite{chenGradNormGradientNormalization2018}, and (2) a modified implementation of AutoPGD~\cite{croceReliableEvaluationAdversarial2020}, for achieving better results than standard PGD, whose effectiveness has been shown to strongly rely on good attack parameters~\cite{mosbachLogitPairingMethods2019,croceScalingRandomizedGradientFree2020}.

\section{Experimental Set-up \& Results}\label{sec:experimental_setup}
All experiments ran on Google Colab; the MNIST experiment described in~\cref{subsec:mnist_experiment} ran on an A100 GPU instance with \qty{40}{\giga\byte} of VRAM, and the drone controller case study described in~\cref{subsec:alsomitra_experiment} ran on a v5e-1 TPU with \qty{47}{\giga\byte} of RAM.
All experiments ran for $100$ epochs and utilised the AdamW~\cite{loshchilovDecoupledWeightDecay2018} optimiser with learning rate \num{1e-3} and weight decay of \num{1e-4}.
The MNIST experiment described in~\cref{subsec:mnist_experiment}, used $20$ PGD iterations and $30$ random restarts; the drone controller experiment described in~\cref{subsec:alsomitra_experiment}, used $50$ PGD iterations and $80$ random restarts. Running all experiments took \qty{1}{\day}~\qty{2}{\hour}~\qty{37}{\minute} to complete.

\paragraph{Evaluation metrics.}
For each experiment, we report results for training without any logical loss (called \emph{baseline}), as well as for training with the DL2 and fuzzy logic differentiable losses.
All experimental results are obtained by measuring the network's performance on the test set using various metrics we will briefly describe in the following.
The full formal definitions can be found in~\cref{eq:pacc,eq:rmse,eq:cacc,eq:csec} in~\cref{sec:appendix}.

\noindent In order to evaluate the network's prediction performance, for classification problems we report the network's \emph{Prediction Accuracy (PAcc)}, defined as the fraction of correct predictions, whereas for regression problems, we make use of the \emph{Root Mean Squared Error (RMSE)} metric, defined as the average difference between the network's predictions and the target values.

Additionally, in order to quantify how well a network satisfies a logical constraint $\phi$, we define two additional evaluation metrics, first proposed in~\cite{casadioNeuralNetworkRobustness2022}.
We first define \emph{Constraint Accuracy (CAcc)} as the fraction of \emph{random samples} that satisfy constraint $\phi$.
In addition to that, let \emph{Constraint Security (CSec)} be the fraction of \emph{adversarial samples} that satisfy constraint $\phi$.

\subsection{Case Study: MNIST}\label{subsec:mnist_experiment}
In order to demonstrate the effectiveness of our approach on a well-known machine learning example, we will train a simple convolutional neural network with ReLU activations to satisfy the standard robustness~\cite{casadioNeuralNetworkRobustness2022} constraint shown in~\cref{eq:standard_robustness} on the MNIST~\cite{lecunGradientbasedLearningApplied1998} data set.
\begin{equation}\label{eq:standard_robustness}
    \forall\vec{x}'\in\mathbb{B}(\vec{x};\epsilon)\ldotp\ \lVert f(\vec{x})-f(\vec{x}')\rVert_{\infty}\le\delta
\end{equation}
Here, the hyper-rectangles that specify the relevant input regions are obtained as the $\epsilon$-cubes around all inputs in the train set.
We chose $\epsilon=0.3$ and $\delta=0.05$.

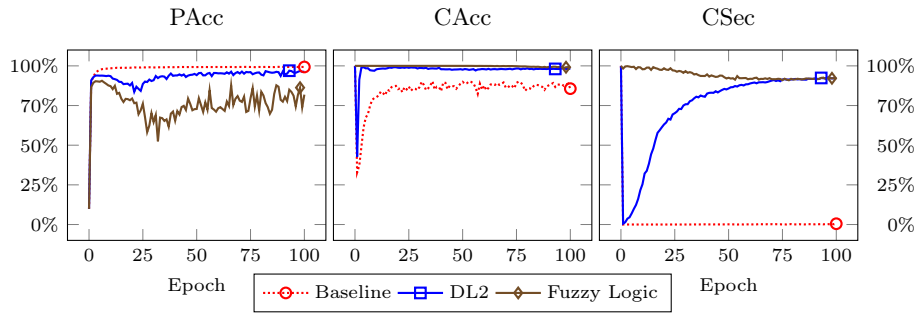
\begin{figure}
    \centering
    \pgfplotsset{
  results/.style={
    width=5cm,
    height=4.1cm,
    ytick distance=0.2,
    xtick={0,25,...,100},
    tick label style={font=\scriptsize},
    xlabel style={font=\scriptsize},
    title style={font=\small},
    ytick={0,0.25,...,1},
    ymin=0, ymax=1,
    yticklabels={0\%,25\%,50\%,70\%,100\%}, 
    legend cell align={left},
    legend style={
      legend columns=3,
      anchor=center,
      align=center,
      font=\scriptsize
    },
    legend image post style={mark indices={}},
    enlargelimits=true,
    xlabel=Epoch,
    every axis plot post/.append style={
      thick,
      table/col sep=comma,
      table/x=Epoch,
    },
  },
  group/results/.style={
    group style={
      group size=3 by 1, 
      ylabels at=edge left,
      x descriptions at=edge bottom,
      horizontal sep=0.1cm,
    }, 
    results,
  },
  baseline/.style={
    mark indices=101, mark=o, red, densely dotted, mark options={solid}
  },
  dl2/.style={
    mark indices=94, mark=square, blue
  },
  fl/.style={
    mark indices=99, mark=diamond, brown!60!black
  }
}

\begin{tikzpicture}[font=\small]
  \begin{groupplot}[group/results]
    \nextgroupplot[title={PAcc},]
\addplot[baseline] table [y=Test-P-Metric] {csv/MNIST/Baseline.csv};
\addplot[dl2] table [y=Test-P-Metric] {csv/MNIST/DL2.csv};
\addplot[fl] table [y=Test-P-Metric] {csv/MNIST/GD.csv};

\coordinate (c1) at (rel axis cs:0,1);
    \nextgroupplot[title={CAcc},
      yticklabels={},
      xlabel={},
    ]
\addplot[baseline] table [y=Test-C-Acc] {csv/MNIST/Baseline.csv};
\addplot[dl2] table [y=Test-C-Acc] {csv/MNIST/DL2.csv};
\addplot[fl] table [y=Test-C-Acc] {csv/MNIST/GD.csv};

\coordinate (c2) at (rel axis cs:0,1);
    \nextgroupplot[title={CSec},
      yticklabel pos=right,
      yticklabel style={anchor=east,xshift=2.7em},
      legend to name=full-legend
    ]
\addplot[baseline] table [y=Test-C-Sec] {csv/MNIST/Baseline.csv};
\addplot[dl2] table [y=Test-C-Sec] {csv/MNIST/DL2.csv};
\addplot[fl] table [y=Test-C-Sec] {csv/MNIST/GD.csv};
\addlegendentry {Baseline};
\addlegendentry {DL2};
\addlegendentry {Fuzzy Logic};

\coordinate (c3) at (rel axis cs:1,1);
  \end{groupplot}
  \coordinate (c4) at ($(c1)!.5!(c3)$);
  \node[below, yshift=0.5cm] at (c4 |- current bounding box.south) {\pgfplotslegendfromname{full-legend}};
\end{tikzpicture}%
    \caption{Training on MNIST without differentiable logics (baseline), with DL2, and with fuzzy logic for the standard robustness property defined in~\cref{eq:standard_robustness}.}
    \label{fig:mnist_results}
\end{figure}

The results are shown in~\cref{fig:mnist_results,tab:mnist_results} and report prediction accuracy (PAcc), constraint satisfaction on random samples (CAcc), and on adversarial samples (CSec), as defined in~\cref{eq:pacc,eq:cacc,eq:csec}.
To avoid prioritising one metric over another, we consider the best result to be the one that maximises their product.
\Cref{tab:mnist_results} reports the best epoch of the last $10$ epochs, which is also marked in~\cref{fig:mnist_results}.
It can be seen that property-driven training leads to significantly improved constraint satisfaction at the expense of prediction accuracy.

\begin{table}
    \centering
    \caption{Training on MNIST without differentiable logics (baseline), with DL2, and with fuzzy logic for the standard robustness property defined in~\cref{eq:standard_robustness}. The best-performing experiment is highlighted in boldface.}
    \label{tab:mnist_results}
    \begin{tblr}
    {
        colspec={Q[l, m, mode=text]S[table-format=3.2, table-auto-round]S[table-format=3.2, table-auto-round]S[table-format=3.2, table-auto-round]},
        row{1}={guard, mode=text, font=\bfseries},
        row{3}={font=\bfseries},
    }
    \toprule
        Logic & PAcc (\%) & CAcc (\%) & CSec (\%) \\
    \midrule
        Baseline & 99.30 & 85.64 & 0.49 \\
        DL2 & 96.96 & 98.08 & 92.38 \\
        Fuzzy Logic & 86.27 & 99.15 & 92.18 \\ \bottomrule
  \end{tblr}
\end{table}

\subsection{Case Study: Neural Network Drone Controller}\label{subsec:alsomitra_experiment}
\begin{figure}
    \centering
    \begin{subfigure}[b]{0.46\textwidth}
        \centering
        \includegraphics[width=.65\linewidth,angle=270,origin=c]{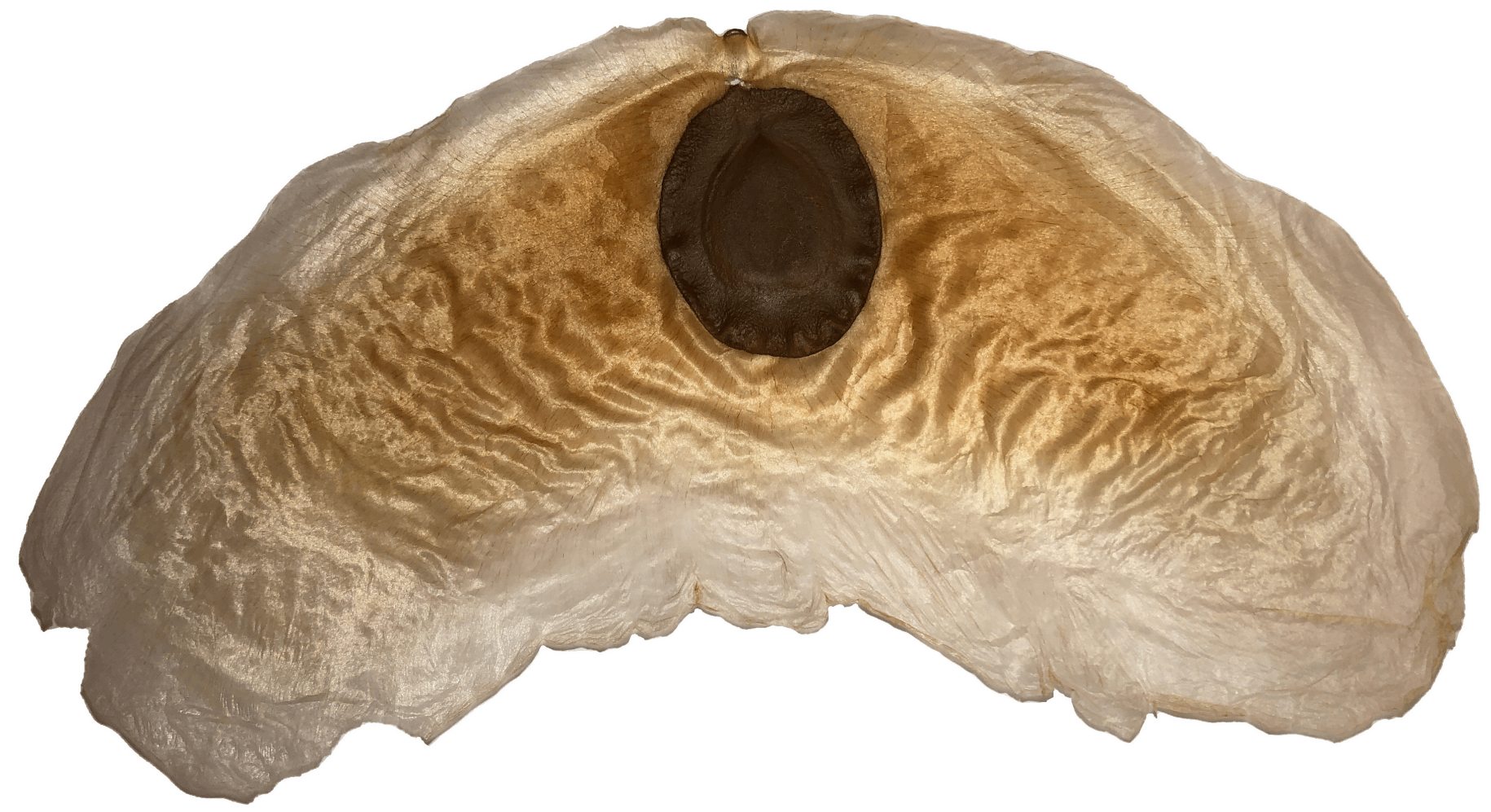}%
        \caption{An \emph{Alsomitra macrocarpa} seed, capable of stable flight over long distances due to its unique shape~\cite{certiniFlightAlsomitraMacrocarpa2023}.}
        \label{fig:alsomitra_seed}
    \end{subfigure}%
    \hfil
    \begin{subfigure}[b]{0.46\textwidth}
        \centering
        \input{figures/alsomitra_control_diagram}%
        \caption{As a control problem, the \emph{Alsomitra}-inspired drone should follow a linear trajectory by actuating its centre of mass.}
        \label{fig:alsomitra_trajectory}
    \end{subfigure}
    \caption{An overview of the \emph{Alsomitra}-inspired drone controller.}
    \label{fig:alsomitra}
\end{figure}

Because the training data for ACAS Xu is not publicly released, we instead consider a neural network controller with similar characteristics as a more sophisticated case study.
Our case study, proposed in~\cite{kesslerNeuralNetworkVerification2025}, revolves around a neural network controller for a gliding drone inspired by the flying seeds of \emph{Alsomitra macrocarpa}, the Javan cucumber, shown in~\cref{fig:alsomitra_seed}.
These flying seeds are capable of stable flight over long distances due to their unique shape~\cite{certiniFlightAlsomitraMacrocarpa2023}.

\noindent The aim of the neural network controller is for the gliding drone to follow a desired trajectory in two-dimensional space (shown in~\cref{fig:alsomitra_trajectory}).
The path of the drone is simulated with a two-dimensional aerodynamic model for falling plates with displaced centre of mass~\cite{liCentreMassLocation2022}, whereby the controller actuates the position of the centre of mass to alter the drone trajectory.
The neural network controller takes as input the six parameters of the dynamic system (local $x$- and $y$-velocities, pitch angular velocity, pitch angle, and global $x$- and $y$-positions) shown with their ranges in~\cref{tab:alsomitra_controller_inputs} in~\cref{sec:appendix} and outputs a value (the displacement of the position of the centre of mass) that can range from \numrange{0.181}{0.193}.
The equation for the desired trajectory is $y=-x$.

\noindent We use a small three-layer feedforward neural network with hidden layers of sizes $64$ and $32$, respectively.
Each layer is followed by a $\text{ReLU}$ activation.

\paragraph{Desired properties.}
In general, we want the drone to follow the trajectory as closely as possible; more specifically, we want it to satisfy the properties defined formally in~\cref{eq:prop1,eq:prop2,eq:prop3,eq:lipschitz_robustness} below.
Note that each property is defined locally for an input $\vec{x}_0$.
\begin{enumerate}
    \item \emph{If the drone is far above the line, the neural network output will always make the drone pitch down:}
    \begin{equation}\label{eq:prop1}
        \phi_1(\vec{x}_0)\colon\quad \forall\vec{x}\ldotp\  \mathsf{farAboveLine}(\vec{x}_0,\vec{x})\implies f(\vec{x})\ge 0.187.
    \end{equation}
    \item \emph{If the drone is close to the line and at an appropriate pitch angle, the neural network output will be intermediate:}
    \begin{equation}\label{eq:prop2}
        \begin{split}
        \phi_2(\vec{x}_0)\colon\quad \forall\vec{x}\ldotp\ &\mathsf{closeToLine}(\vec{x}_0,\vec{x})\\\wedge\ &\mathsf{appropriatePitchAngle}(\vec{x})\implies 0.184\le f(\vec{x})\le 0.19.
        \end{split}
    \end{equation}
    \item \emph{If the drone is above and close to the line, pitching down quickly and moving fast, the neural network output will always make the drone pitch up:}
    \begin{equation}\label{eq:prop3}
        \begin{split}
            \phi_3(\vec{x}_0)\colon\quad \forall\vec{x}\ldotp\ &\mathsf{aboveAndCloseToLine}(\vec{x}_0,\vec{x})\\ \wedge\  &\mathsf{pitchingDownQuickly}(\vec{x})\\ \wedge\  &\mathsf{movingFast}(\vec{x}) \implies f(\vec{x})\le 0.187.
        \end{split}
    \end{equation}
    \item Lastly, as neural networks with ReLU activations are Lipschitz-continuous~\cite{szegedyIntriguingPropertiesNeural2014}, and \emph{Lipschitz robustness} is the strongest form of robustness~\cite{casadioNeuralNetworkRobustness2022}, we are particularly interested for the network to satisfy the property:
    \begin{equation}\label{eq:lipschitz_robustness}
        \phi_4(\vec{x}_0)\colon\quad \forall\vec{x}\ldotp\ \mathsf{closeToLine(\vec{x}_0,\vec{x})}\implies\lVert f(\vec{x}_0)-f(\vec{x})\rVert_2\le L\lVert\vec{x}_0-\vec{x}\rVert_2.
    \end{equation}
    Note that a smaller value of $L$ means a smoother output, and therefore a more robust network.
    We use a value of $L=0.3$ in our experiment.
\end{enumerate}
In the above property definitions, we make use of certain predicates to present the desired properties as clearly as possible.
These predicates are defined in~\cref{eq:farAboveLine,eq:closeToLine,eq:appropriatePitchAngle,eq:aboveAndCloseToLine,eq:pitchingDownQuickly,eq:movingFast} in~\cref{sec:appendix}.

The results of training networks to satisfy the desired properties are shown in~\cref{tab:alsomitra_results} and confirm that property-driven training leads to considerably improved constraint satisfaction.
Again, we note that this increase in constraint satisfaction comes at the expense of regression performance, visualised in~\cref{fig:regression_performance} in~\cref{sec:appendix} for the Lipschitz robustness property $\phi_4$.

\begin{table}
    \centering
    \caption{Results of training neural networks without differentiable logics (baseline) or various differentiable logics, and evaluating their regression performance (RMSE), constraint satisfaction on random samples (CAcc) and on adversarial samples (CSec) for the properties $\phi_1$ to $\phi_4$ defined in~\cref{eq:prop1,eq:prop2,eq:prop3,eq:lipschitz_robustness}. For each property, the best performing experiment is highlighted in bold face.}
    \label{tab:alsomitra_results}
    \begin{tblr}
    {
        colspec={Q[c, m, mode=math]Q[l, m, mode=text]S[scientific-notation=true, round-mode=places, round-precision=4, table-format=1.4e-1]S[table-format=3.2, table-auto-round]S[table-format=3.2, table-auto-round]},
        row{1}={guard, mode=text, font=\bfseries},
        row{3,6,12,16}={font=\bfseries},
    }
        \toprule
            Property & Logic & RMSE & CAcc (\%) & CSec (\%) \\
        \midrule
            \SetCell[r=3, c=1]{c} \phi_1 & Baseline & 0.0003611086576711 & 96.875 & 34.375 \\
            & DL2 & 0.0006612562574446 & 100.0 & 98.4375 \\
            & Fuzzy logic & 0.0006887059425935 & 100.0 & 98.4375 \\ \midrule
            \SetCell[r=6, c=1]{c} \phi_2 & Baseline & 0.0003611086576711 & 23.4375 & 0.0 \\
            & DL2 & 0.0010368227958679 & 100.0 & 95.3125 \\
            & Gödel logic & 0.0010778681607916 & 100.0 & 82.8125 \\
            & \L ukasiewicz logic & 0.0010777524439617 & 100.0 & 82.8125 \\
            & Reichenbach logic & 0.0010778681607916 & 100.0 & 82.8125 \\
            & Yager logic & 0.0011235475540161 & 100.0 & 85.9375 \\\midrule
            \SetCell[r=3, c=1]{c} \phi_3 & Baseline & 0.0003611086576711 & 0.0 & 0.0 \\
            & DL2 & 0.0012287280987948 & 100.0 & 95.3125 \\
            & Fuzzy logic & 0.0011632416862994 & 100.0 & 92.1875 \\ \midrule
            \SetCell[r=3, c=1]{c} \phi_4 & Baseline & 0.0003611086576711 & 57.8125 & 0.0 \\
            & DL2 & 0.001849079853855 & 100.0 & 71.875 \\
            & Fuzzy logic & 0.0021656022872775 & 100.0 & 93.75 \\
        \bottomrule
    \end{tblr}
\end{table}

\section{Discussion}\label{sec:discussion}
We have described a generalised framework for property-driven machine learning and evaluated an implementation on two case studies; a standard classification benchmark on MNIST in~\cref{subsec:mnist_experiment}, and a regression case study in~\cref{subsec:alsomitra_experiment}.
In both cases, constraint satisfaction could significantly be improved via property-driven training compared to models trained in a standard manner, however at the expense of prediction performance (as initially reported in~\cite{tsiprasRobustnessMayBe2018}).

\subsection{Related Work}
\paragraph{Property-driven training.}
Some of the ideas presented in this paper were already present in DL2~\cite{fischerDL2TrainingQuerying2019}.
However, while arbitrary box constraints (i.e. hyper-rectangles) are mentioned, no rationale is provided explaining why these might prove to be useful, and the experimental evaluation is limited to $\epsilon$-balls.

Apart from DL2, various other approaches have explored integrating logical constraints into the training process:
Semantic-based Regularisation (SBR)~\cite{diligentiSemanticbasedRegularizationLearning2017a} encodes first order logic rules as additional loss terms.
Logic Tensor Networks~\cite{serafiniLogicTensorNetworks2016} adopt a framework called Real Logic, where symbols and predicates are grounded in real-valued vector spaces such that logical constraints can be integrated directly into the neural network's architecture.
Furthermore, probabilistic differentiable logics such as DeepProbLog~\cite{manhaeveDeepProbLogNeuralProbabilistic2018} and Semantic Loss~\cite{xuSemanticLossFunction2018} combine logical reasoning with probabilistic inference.

\paragraph{Frameworks for property-driven training.}
Several frameworks aim to provide a unified view of property-driven training.
PYLON~\cite{ahmedPYLONPyTorchFramework2022a} is a Python framework for property-driven training, however, its scope is limited to output-level constraints, without employing PGD or other methods to find adversarial examples that violate the constraints.
Similarly, ULLER~\cite{vankriekenULLERUnifiedLanguage2024} is a unifying language for neuro-symbolic AI that allows for fuzzy, classical, and probabilistic semantics.
An implementation does not yet exist and is left for future work.

\paragraph{Architectures that guarantee constraint satisfaction.}
Giunchiglia et al.~\cite{giunchigliaDeepLearningLogical2022} provide a survey of property-driven machine learning, including specialised neural network architectures to obtain hard guarantees of constraint satisfaction.
Recent efforts have focused on integrating logical constraints in a neural network layer such that the network is guaranteed to satisfy the constraints by design.
For example, CCN+~\cite{giunchigliaCCNNeurosymbolicFramework2024} integrates constraints into the output layer of the network, while the Semantic Probabilistic Layer (SPL)~\cite{ahmedSemanticProbabilisticLayers2022} leverages probabilistic circuits to enable exact probabilistic inference with logical reasoning.
More recent work has focused on non-convex constraints: the Disjunctive Refinement Layer (DRL)~\cite{stoianConvexityAssumptionRealistic2024} supports non-convex and disconnected spaces, and the Probabilistic Algebraic Layer (PAL)~\cite{kurscheidtProbabilisticNeurosymbolicLayer2025} guarantees satisfaction of non-convex algebraic constraints.

\subsection{Conclusion}
We present a practical framework for property-driven machine learning that incorporates logical specifications directly into the training process, guiding neural networks to learn to satisfy constraints.
This approach lays the foundation for effective formal verification of neural networks and marks a step towards obtaining correct-by-construction neural networks.

This work can also be seen as a stepping stone towards a more general compilation procedure from logical properties to machine learning optimisation objectives, announced as a target of the Vehicle~\cite{daggittVehicleInterfacingNeural2022,daggittCompilingHigherOrderSpecifications2023,daggittVehicleTutorialNeural2023,daggittVehicleBridgingEmbedding2024,daggittEfficientCompilationExpressive2024} language\footnote{We note that at the time of submitting this paper, Vehicle's algorithm is not finalised and thus we cannot make a technical comparison with the results presented here.}.

\paragraph{Limitations and future work.}
While they are computationally effective, hyper-rectangles are not the most precise geometric representation of an input region.
For example, the properties for the Alsomitra drone controller as originally proposed in~\cite{kesslerNeuralNetworkVerification2025} involve bounds on components of the input vector that depend on its other components, e.g. the bounds on the $y$-component of the input vector $\vec{x}$ depend on the value of its own $x$-component.
These bounds give rise to a half-space, and they cannot be represented as a hyper-rectangle that requires constant lower and upper bounds for each dimension.
Casadio et al.~\cite{casadioNLPVerificationGeneral2025a} note that a convex hull would capture input regions the most accurately, however, constructing this would be computationally expensive.

On the other hand, loss-based methods provide (almost) no formal guarantees, as confirmed in~\cite{flinkowComparingDifferentiableLogics2025}.
Worth exploring in future work is to investigate whether ideas from certified training~\cite{wongProvableDefensesAdversarial2018,wongScalingProvableAdversarial2018,mirmanDifferentiableAbstractInterpretation2018,raghunathanCertifiedDefensesAdversarial2018a} can be integrated into this framework.

\noindent An interesting subject for further study is to define in more precise terms the fragment of FOL that can covered by the admissible verification properties that correspond to optimisation objectives.
The property $\forall\vec{x}\in\mathbb{R}^m\ldotp \mathcal{P}(\vec{x})\implies \mathcal{Q}(f(\vec{x}))$ clearly has a clausal form, and there is a long history of studying first-order Horn clauses and first-order hereditary Harrop clauses as optimal fragments of FOL for automated reasoning and verification~\cite{millerProgrammingHigherOrderLogic2012,bjornerHornClauseSolvers2015}.

\begin{credits}
\subsubsection{\ackname}
Flinkow and Monahan acknowledge the support of Taighde Éireann – Research Ireland grant number 20/FFP-P/8853.
Komendantskaya acknowledges the partial support of the EPSRC grant AISEC: AI Secure and Explainable by Construction (EP/T026960/1),
and support of ARIA: Mathematics for Safe AI grant.
Kessler is funded by the EPCRS CDT ``Edinburgh Centre for Robotics''.
Casadio acknowledges a James Watt PhD Scholarship from Heriot-Watt University. 

\subsubsection{\discintname}
The authors have no competing interests to declare that are relevant to the content of this article.
\end{credits}

\appendix
\section{Appendix}\label{sec:appendix}

\subsection{Evaluation Metrics}
All experimental results are obtained by measuring the network's performance on the test set $\mathcal{T}=\{(\vec{x}_i,\vec{y}_i)\}_{i=1}^n$ using various metrics we will define in the following.
The test set consists of pairs of input data $\vec{x}_i$ and target data $\vec{y}_i$ (i.e. the true label for classification, or the target value for regression).

Let $[P]$ denote the Iverson bracket, returning $1$ if $P$ is true, and $0$ otherwise, and let $\dl{\mathcal{Q}(f(\vec{x}))}{BL}$ denote the standard Boolean logic interpretation of $\mathcal{Q}(f(\vec{x}))$, used to evaluate whether the constraint is satisfied.

In order to evaluate the network's prediction performance, for classification problems we report the network's \emph{Prediction Accuracy (PAcc)}, defined as the fraction of correct predictions:
\begin{equation}\label{eq:pacc}
    \text{PAcc}:=\frac{1}{|\mathcal{T}|}\sum\limits_{(\vec{x},\vec{y})\in\mathcal{T}}[f(\vec{x})=\vec{y}],
\end{equation}
whereas for regression problems, we make use of the \emph{Root Mean Squared Error (RMSE)} metric, defined as the average difference between the network's predictions and the target values:
\begin{equation}\label{eq:rmse}
    \text{RMSE}:=\sqrt{\frac{1}{|\mathcal{T}|}\sum\limits_{(\vec{x},\vec{y})\in\mathcal{T}}(f(\vec{x})-\vec{y})^2}.
\end{equation}
Additionally, in order to quantify how well a network satisfies a logical constraint $\phi$, we define two additional evaluation metrics, first proposed in~\cite{casadioNeuralNetworkRobustness2022}.
We define \emph{Constraint Accuracy (CAcc)} as the fraction of \emph{random samples} $\vec{x}_{\text{rnd}}$ that satisfy constraint $\phi$:
\begin{equation}\label{eq:cacc}
    \text{CAcc}:=\frac{1}{|\mathcal{T}|}\sum\limits_{(\vec{x},\vec{y})\in\mathcal{T}}[\dl{\mathcal{Q}(f(\vec{x}))}{BL}(\vec{x}; \vec{x}_{\text{rnd}};\vec{y})],
\end{equation}
where, for each $(\vec{x},\vec{y})\in\mathcal{T}$, a fresh random sample $\vec{x}_{\text{rnd}}$ is drawn from $\hrect{\mathcal{P}(\vec{x})}$.

In addition to that, let \emph{Constraint Security (CSec)} be the fraction of \emph{adversarial samples} $\vec{x}_{\text{adv}}$ within that satisfy constraint $\phi$:
\begin{equation}\label{eq:csec}
    \text{CSec}:=\frac{1}{|\mathcal{T}|}\sum\limits_{(\vec{x},\vec{y})\in\mathcal{T}}[\dl{\mathcal{Q}(f(\vec{x}))}{BL}(\vec{x}; \vec{x}_{\text{adv}};\vec{y})],
\end{equation}
where, for each $(\vec{x},\vec{y})\in\mathcal{T}$, a fresh adversarial sample $\vec{x}_{\text{adv}}$ within $\hrect{\mathcal{P}(\vec{x})}$ is obtained using PGD.

\subsection{Alsomitra Drone Controller Inputs}
\Cref{tab:alsomitra_controller_inputs} describes the $6$ inputs to the Alsomitra drone controller with their intended meaning, unit, and lower and upper bounds.
\begin{table}
    \sisetup{table-format=+2.2, table-auto-round}
    \centering
    \caption{The Alsomitra drone controller inputs and their ranges.}
    \label{tab:alsomitra_controller_inputs}
    \begin{tblr}
    {
        colspec={Q[c, m, mode=math]lQ[c, m, mode=text]SS},
        row{1}={guard, mode=text, font=\bfseries},
        row{2}={guard, mode=text}
    }
        \toprule
            \SetCell[r=2,c=1]{l} Parameter & \SetCell[r=2,c=1]{l} Description & \SetCell[r=2,c=1]{l} Unit & \SetCell[r=1, c=2]{c} Range \\ \cline{4-5}
            & & & Lower & Upper \\
        \midrule
            v_{x'} & local $x$-velocity & \unit{\meter\per\second} & 0.967568147 & 3.489893068 \\
            v_{y'} & local $y$-velocity & \unit{\meter\per\second} & -0.607397104 & -0.043021391 \\
            \omega & {pitch angular velocity} & \unit{\radian} & -0.356972794 & 0.049180223 \\
            \theta & {pitch angle} & \unit{\radian} & -0.96847853 & -0.068002517 \\
            x & global $x$-position & \unit{\meter} & 0.482420778 & 41.714716998 \\
            y & global $y$-position & \unit{\meter} & -41.68140527 & 4.19723352 \\
        \bottomrule
    \end{tblr}
\end{table}

\subsection{Alsomitra Drone Controller Property Definitions}
First, let $\vec{x}=[v_{x'}, v_{y'}, \omega, \theta, x, y]$ denote an input vector of the Alsomitra controller neural network, consisting of the six system parameters.
We then use $\omega^{(\vec{x})}$ to refer to the $\omega$ component of $\vec{x}$ (similarly for its other components).
Then, we define the predicates used in the Alsomitra drone controller properties in~\cref{eq:prop1,eq:prop2,eq:prop3} as shown below:

\begin{align}
    &\mathsf{farAboveLine}(\vec{x}_0,\vec{x})&&:=\quad y^{(\vec{x})}\ge 2 - x^{(\vec{x}_0)} \label{eq:farAboveLine} \\
    &\mathsf{closeToLine}(\vec{x}_0,\vec{x})&&:=\quad -2 - x^{(\vec{x}_0)} \le y^{(\vec{x})} \le 2 - x^{(\vec{x}_0)} \label{eq:closeToLine} \\
    &\mathsf{appropriatePitchAngle}(\vec{x})&&:=\quad -0.786 \le \theta^{(\vec{x})} \le -0.747 \label{eq:appropriatePitchAngle} \\
    &\mathsf{aboveAndCloseToLine}(\vec{x}_0,\vec{x})&&:=\quad -x^{(\vec{x}_0)} \le y^{(\vec{x})} \le 2 - x^{(\vec{x}_0)} \label{eq:aboveAndCloseToLine} \\
    &\mathsf{pitchingDownQuickly}(\vec{x})&&:=\quad \omega^{(\vec{x})} \le -0.12 \label{eq:pitchingDownQuickly} \\
    &\mathsf{movingFast}(\vec{x})&&:=\quad v_{x'}^{(\vec{x})} \le -0.3 \label{eq:movingFast}
\end{align}

\subsection{Alsomitra Drone Controller Experimental Results}
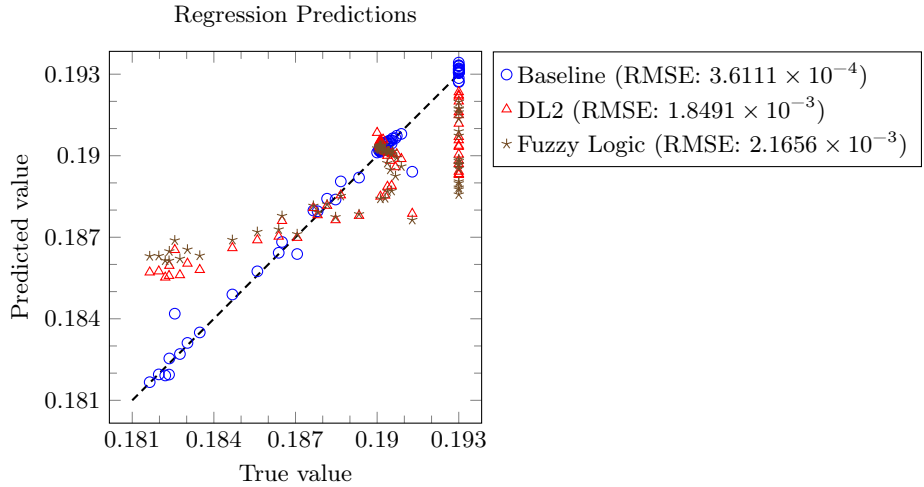
\begin{figure}
    \centering
        \begin{tikzpicture}
        \begin{axis}[
            xlabel=True value,
            ylabel=Predicted value,
            title=Regression Predictions,
            width=6.5cm,
            height=6.5cm,
            xmin=0.181,
            xmax=0.193,
            ymin=0.181,
            ymax=0.193,
            minor tick num=2,
            xtick={0.181,0.184,0.187,0.190,0.193},
            ytick={0.181,0.184,0.187,0.190,0.193},
            enlargelimits=0.07,
            legend pos=outer north east,
            legend cell align={left},
            tick label style={/pgf/number format/precision=3},
        ]
        \addplot[only marks, blue, mark=o] table[col sep=comma, x=target, y=predicted] {csv/Alsomitra/Baseline-RegressionPlot.csv};
        \addlegendentry{Baseline (RMSE: \num[scientific-notation=true, round-mode=places, round-precision=4]{0.0003611086576711})};
        
        \addplot[only marks, red, mark=triangle] table[col sep=comma, x=target, y=predicted] {csv/Alsomitra/DL2-RegressionPlot.csv};
        \addlegendentry{DL2 (RMSE: \num[scientific-notation=true, round-mode=places, round-precision=4]{0.001849079853855})};
        
        \addplot[only marks, brown!60!black, mark=star] table[col sep=comma, x=target, y=predicted] {csv/Alsomitra/GD-RegressionPlot.csv};
        \addlegendentry{Fuzzy Logic (RMSE: \num[scientific-notation=true, round-mode=places, round-precision=4]{0.0021656022872775})};
        
        \addplot[
            domain=0.181:0.193,
            black,
            thick,
            densely dashed
        ] {x};
        
        \end{axis}
    \end{tikzpicture}
    \caption{A scatter plot of predicted vs. true values for all elements in the test set for networks trained to satisfy the Lipschitz robustness constraint $\phi_4$ defined in~\cref{eq:lipschitz_robustness}. Regression performance is reduced for networks obtained via property-driven training compared to the baseline.}
    \label{fig:regression_performance}
\end{figure}

\Cref{fig:regression_performance} visualises the decrease in regression performance that occurs with property-driven training on one of the properties of the Alsomitra controller.

\bibliographystyle{splncs04}
\bibliography{references}

\end{document}